\pdfoutput=1
\documentclass[11pt]{article}

\usepackage[final]{acl}

\usepackage{times}
\usepackage{latexsym}
\usepackage{graphicx}
\usepackage[inkscapelatex=false]{svg}

\usepackage[T1]{fontenc}
\usepackage[utf8]{inputenc}

\usepackage{microtype}

\usepackage{inconsolata}
\title{Eigenpruning: an Interpretability-Inspired PEFT Method}

\author{
Tomás Vergara-Browne\textsuperscript{123}, Álvaro Soto\textsuperscript{12}, Akiko Aizawa\textsuperscript{3} \\
\textsuperscript{1}Pontificia Universidad Católica de Chile \\
\textsuperscript{2}Centro Nacional de Inteligencia Artificial, Chile \\
\textsuperscript{3}National Institute of Informatics, Japan
}

\begin{document}
\maketitle
\begin{abstract}
We introduce \textit{eigenpruning}, a method that removes singular values from weight matrices in an LLM to improve its performance in a particular task. This method is inspired by interpretability methods designed to automatically find subnetworks of a model which solve a specific task. In our tests, the pruned model outperforms the original model by a large margin, while only requiring minimal computation to prune the weight matrices. In the case of a small synthetic task in integer multiplication, the Phi-2 model can improve its accuracy in the test set from 13.75\% to 97.50\%. Interestingly, these results seem to indicate the existence of a computation path that can solve the task very effectively, but it was not being used by the original model. Finally, we publicly release our implementation\footnote{\href{https://github.com/tvergara/eigenpruning}{https://github.com/tvergara/eigenpruning}}.
\end{abstract}

\section{Introduction}

Recently there has been a great interest in understanding, or \textit{reverse-engineering} circuits of LLMs \cite{elhage2021mathematical, nanda2023progress}. Most of these works have been manually detecting interpetable circuits in LLMs \cite{nanda2023progress, wang2022interpretability, hanna2024does}. But manually \textit{reverse-engineering} the LLMs of today is infeasable, due to their size, making automatic approaches important for the future of the field \cite{rauker2023toward}. There has been some early work on this topic \cite{conmy2023towards, syed2023attribution}, which has had some success into finding interpretable circuits.

We are inspired by the "\textit{competing subnetworks}" theory \cite{merrill2023tale}, where model training is understood as a competition of multiple algorithms to solve a task. These subnetworks may generalize very differently \cite{bhaskar2024heuristic}, we theorize that there exist \textit{weaker} subneworks, that could be pruned to enhance performance.

By applying a variation of Attribution Patching \cite{syed2023attribution}, an automatic method to discover circuits, we remove singular values from some matrices of an LLM to achieve better performance than the original model, both in real NLP tasks and synthetic tasks. We call our method \textit{eigenpruning}.

\section{Related Work}
To the best of our knowledge, there exists 2 automated circuit discovery approaches. They view models as computational graphs, and circuits are defined as subgraphs with distinct functionality, as in \cite{wang2022interpretability}. These approaches are:
\begin{itemize}
    \item \textbf{ACDC} \cite{conmy2023towards}: Starting from the output node, ACDC iterates over each node in the computational graph aiming to remove as many edges as possible without affecting performance. Unfortunately, this approach does not scale favourably with the number of nodes and edges in the graph, which makes it infeasible for extremely large models.
    \item \textbf{Attribution patching} \cite{syed2023attribution}: In contrast to ACDC, Attribution Patching performs a linear approximation of the effect of removing an edge of the graph. This is much more efficient in terms of computation, as it requires a constant number of forward and backward passes with respect to the size of the computational graph.
\end{itemize}
\begin{figure*}[]
    \centering
    \includegraphics[width=2\columnwidth]{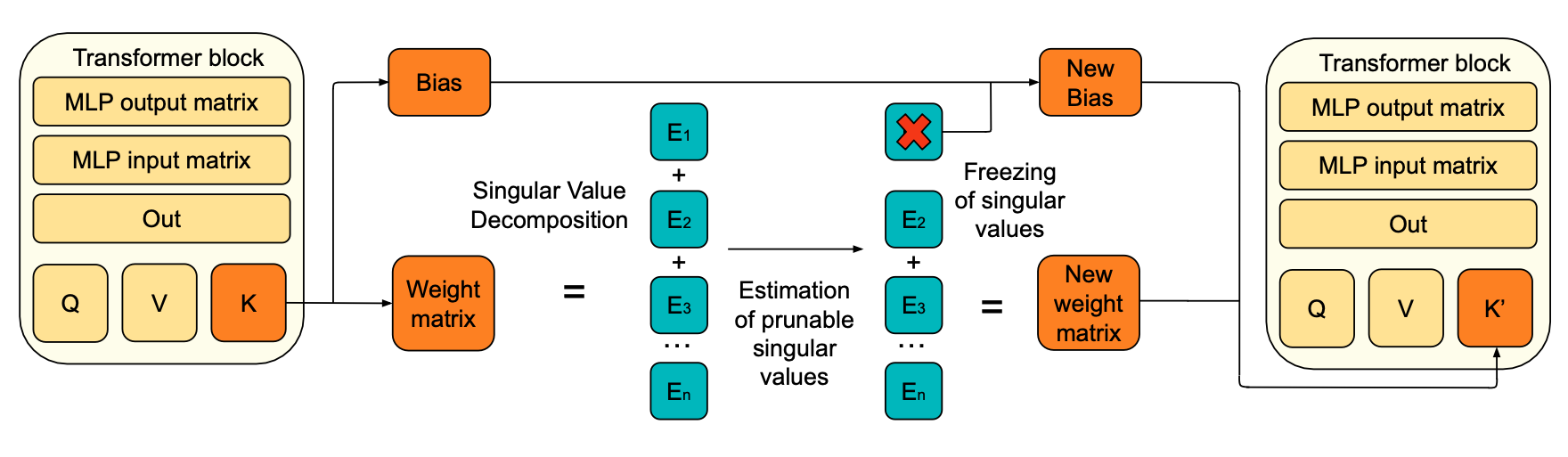}
    \caption{High-level diagram of the \textit{eigenpruning} method.}
    \label{fig:diagram}
\end{figure*}

These previous approaches provide valueable insights:
\begin{itemize}
    \item Both approaches, despite being general to any definition of the computational graph, use \textit{"big"} nodes (attention heads and MLP layers) in their definitions. We do no think that the computations in an LLM are well decomposed in such components, as in \cite{geva2022transformer}, where they found that decomposing feed forward layers outputs into multiple vectors made them more interpretable.
    \item In Attribution Patching, the authors straightforwardly remove edges from the computational graph. We argue that such method to remove edges can produce activations far away from the original model's activation distribution.
    \item Other work has shown that using attribution methods to prune attention heads can improve the accuracy in certain tasks \cite{hao2021self}.
\end{itemize}
Based on these insights, we propose \textit{eigenpruning}, a compute-efficient method to remove singular values from matrices in a model, which can considerably improve the performance of the model in specific tasks.

\section{Eigenpruning}
\label{sec:methods}

A diagram of the method can be found in Figure \ref{fig:diagram}. We first need to manually choose a hyperparameter $M$, that is the subset of the weight matrices inside a model that we are going to use (for example, all the key matrices from transformer blocks \cite{vaswani2017attention}).

For each matrix $A \in M$, we compute its singular value decomposition as:
$$ A = USV$$

Where $S$ is a rectangular diagonal matrix containing the singular values of $A$. Let $n$ be the number of singular values of $A$, from $USV$ we can compute $n$ rank-one matrices $E_i$ corresponding to each singular value.
$$ A = E_1 + E_2 + ... + E_n$$

Very much inspired by Attribution Patching \cite{syed2023attribution}, we use a linear approximation to estimate how intervening the effect of $E_i$ would affect the loss. Let $x$ and $x'$ be two activations corresponding to two separate examples in the training set, which will get multiplied by $A$, then we define this approximation as:
$$ \Delta L_i^x = \frac{\delta L^x}{E_ix} E_i(x' - x)$$

Where $\frac{\delta L^x}{E_ix}$ is the gradient of the loss in example $x$ in the $E_ix$ value.

This is an approximation on the effect of replacing the activation $x$ with $x'$. Considering that we can draw many pairs $(x, x')$ from our training set, we take the maximum from these values as the final $ \Delta L_i $. Our intuition is that $E_i$ values which are part of weak computation paths will result in negative values of $ \Delta L_i $, which indicate that the singular value is degrading the performance.

\begin{table*}
\centering
\begin{tabular}{lccccc}
\hline
\textbf{Model} & \textbf{CB} (\%) & \textbf{COPA} (\%) & \textbf{RTE} (\%) & \textbf{INT-SUM} (\%)  & \textbf{INT-MULT} (\%) \\
\hline
    Phi-2 & 42.86$\rightarrow$51.79 & 78.00$\rightarrow$84.00 & 42.96$\rightarrow$44.04 & 1.45 $\rightarrow$ 46.10 & 13.75 $\rightarrow$ 97.50\\ 

GPT-2 & 10.71$\rightarrow$12.50 & 55.00$\rightarrow$55.00 & 00.36$\rightarrow$1.44 & 3.00 $\rightarrow$ 3.20 & 13.75 $\rightarrow$ 15.00 \\
FT-GPT-2 & 41.07$\rightarrow$50.00 & 55.00$\rightarrow$ 55.00 & 47.29$\rightarrow$52.71& 4.00 $\rightarrow$ 4.00 & 0.00 $\rightarrow$ 0.00\\ 
\hline
\end{tabular}
\caption{Summary of accuracy improvements in the test set in different benchmarks before and after the \textit{eigenpruning}.}
\label{tab:results}
\end{table*}

We now set the hyperparameter $p$, and \textit{freeze} the lowest $p$ portion of singular values. Freezing a singular value $E_i$ is fixing its effect, to make it input-independent. More formally, if you let the $b$ be the bias associated to the $A$, we then update the values as:
$$A' = A - E_i \textnormal{ \ \ \  and \ \ \ } b' = b + E_i\bar x$$
Where $\bar x$ is the average of $x$ activations that were multiplied by $A$ in the training set. This \textit{freezes} the effect the singular values to just the average of the $x$ activations. One important consideration is that the activations $x$ have a token dimension, and we let $b'$ inherit that same dimension. This makes the new bias of the layer to be token-index-dependent.

\section{Experiments}

We tested our approach in two synthetic datasets (integer addition and integer multiplication), which details can be found in the appendix \ref{sec:synthetic-datasets}, and also 3 tasks from the  SuperGLUE benchmark \cite{wang2019superglue}: CB \cite{de2019commitmentbank}, COPA \cite{roemmele2011choice} and RTE \cite{wang2018glue}. Details on the prompting for these tasks can be found in the appendix \ref{sec:nlp-datasets}.

We used 2 open-source LLMs: GPT2-XL \cite{radford2019language}, Phi-2 \cite{javaheripi2023phi}. For GPT2-XL, we also finetune it to each task before applying our method to check whether there is an effect of combining finetuning with \textit{eigenpruning}. Details can be found in Appendix \ref{sec:finetuning}.

In our tests, using the set of matrices $M$ as all the key matrices transformer blocks provided the best results, and we tested different values of the portion of singular values to prune $p$ from the set of $\{ 0.01, 0.05, 0.10, 0.30$\}.

\section{Results}

Table 1 summarizes the improvement that \textit{eigenprunning} models has in each of the datasets used. We draw some insights from these results.

First, \textit{eigenpruning} a model improves the accuracy in each task. This result is much more noticeable in Phi-2 than in GPT-2. This improvement in performance can be very significant, such as in INT-SUM and INT-MULT, although the results should be taken with a grain of salt, as these datasets are small and very well suited for the method (Appendix \ref{sec:synthetic-datasets}).

More interestingly, \textit{eigenpruning} does increase the accuracy in our NLP tasks. For example, the 6\% improvement of Phi-2 in COPA is promising in terms of the practical application of the method.

\begin{table}
\centering
\small
\begin{tabular}{p{0.9cm}|l|ccccc}
\hline
\textbf{Model} & \textbf{Dataset} & Base & \textbf{0.01} & \textbf{0.05} & \textbf{0.10} & \textbf{0.30} \\
\hline
 & INT-SUM & 1.45 & 0.85 & \textbf{46.10} & 7.85 & 8.30 \\
                       & INT-MULT  & 13.75 & 13.75 & \textbf{97.50} & 65.00 & 73.75 \\
       Phi-2                & CB & 42.86 & \textbf{51.79} & 0.00 & 0.00 & 0.00 \\
                       & COPA & 78.00 & \textbf{84.00} & 0.00 & 0.00 & 0.00 \\
                       & RTE & 42.96 & \textbf{44.04} & 0.00 & 0.00 & 0.00 \\
\hline
 & INT-SUM & 3.00 & 2.95 & \textbf{3.20} & 3.00 & 2.45 \\
                       & INT-MULT  & 13.75 & 13.75 & \textbf{15.00} & \textbf{15.00} & \textbf{15.00} \\
            GPT-2           & CB & 10.71 & 10.71 & 10.71 & \textbf{12.50} & 0.00 \\
                       & COPA & \textbf{55.00} & 45.00 & 54.00 & \textbf{55.00} & 46.00 \\
                       & RTE & 0.36 & 0.36 & 0.36 & \textbf{1.44} & 0.00 \\
\hline
 & INT-SUM & \textbf{4.00} & \textbf{4.00} & \textbf{4.00} & \textbf{4.00} & \textbf{4.00} \\
          FT-                & INT-MULT & 0.00 & 0.00 & 0.00 & 0.00 & 0.00 \\
     GPT-2                     & CB & 41.07 & 41.07 & \textbf{50.00} & 41.07 & 41.07 \\
                              & COPA & 55.00 & \textbf{55.00} & \textbf{55.00} & 45.00 & \textbf{55.00} \\
                          & RTE & 47.29 & \textbf{52.71} & \textbf{52.71} & \textbf{52.71} & \textbf{52.71} \\
\hline
\end{tabular}
\caption{Accuracy in the test set in all tasks after \textit{eigenpruning}. $p$ values used are denoted at the top. Top results by model are in bold.}
\label{tab:full-results}
\end{table}

\section{Discussion}
These results are promising, both in terms of performance and of our understanding of how LLMs work.

First, \textit{eigenpruning} models has improved the accuracy of Phi-2 our synthetic tasks, and also (in a lesser extent) in real NLP tasks. This indicates the potential of \textit{eigenpruning} as a method to increase the performance of LLMs.

Some of these results are very surprising at first sight. For example, it is not intuitive that Phi-2 cannot properly perform integer multiplication in its base form, but when pruned of particular singular values of the key matrices in its transformer blocks, can achieve very impressive performance. This makes us believe that the model has computation paths that are capable of solving the task, but it is not properly using them.

\section{Limitations}
This is work in progress, and many limitations are present. 

First, we have not tried sufficient models to be able to understand if the potential of \textit{eigenpruning} is truly generalizable to all LLMs. For example, GPT-2 had much less of a benefit from \textit{eigenpruning} than Phi-2. It might be the case that \textit{eigenpruning} just needs more capable models to be useful, but we do not currently have the data to make that claim.

Second, the optimal $p$ value for each combination of task/model is currently done as a brute force search over several candidate values, which is far from being ideal.

Third, we have not enough data to understand the effect of finetuning with \textit{eigenpruning}. We only tested finetuning GPT-2, but we need to test more models and finetune each one to actually understand its effects.

Fourth, our work needs comparison to other parameter efficient finetuning methods, such as LoRA \cite{hu2021lora}, and potentially also with model editing approaches, such as \cite{mitchell2021fast}.

Finally, the results on synthetic datasets should be taken with a grain of salt, such as explained in appendix \ref{sec:synthetic-datasets}.

\section{Acknowledgements}
We appreciate the reviewers' feedback on this version of our work. We received suggestions for related work that we were previously not aware about. We believe these might be very influential in our future work in \textit{eigenpruning}.

Also, we would like to thank LatinX for the opportunity to present our work. They are doing amazing work into providing opportunities for Latinx individuals along the world.

Finally, we would like to thank Aizawa-lab students for their feedback in previous versions of our diagrams, and overall suggestions over the course of this research.
\bibliography{acl_latex}

\appendix

\section{Details on Synthetic Datasets}
\label{sec:synthetic-datasets}
The integer addition was created by iterating for all the combinations of adding two integers (such as "\verb|2 + 2 =|") with integers from 0 to 99, and let the model complete the last token. This results in 10k samples, which are split into a train and test set in 80\% and 20\% respectively.

The integer multiplication was created very similarly, with the difference of using 19 as the top value instead of 99. This results in 400 samples, which are also then randomly split in train and test set in 80\% and 20\% proportion. The reason behind this choice is that some integers above 400 may get split into different tokens by the tokenizers of the model, which makes the evaluation slightly harder to implement (but will be done in future work).

We know the size of the multiplication dataset is extremely small to actually generalize the results, so that is an important limitation of our results in this task.

Also, in both datasets, the distribution of tokens in these datasets are "perfectly aligned", in the sense that for a particular index, it is consistent whether that index token corresponds to the operation token, the equal sign token, or an integer token. We believe this to be very important into explaining the performance of our method in the tasks, due to the added token dimension in the updated bias in each weight matrix, as explained in Section \ref{sec:methods}.

\section{Details on NLP Datasets}
\label{sec:nlp-datasets}
These datasets are prompted with a manually designed prompt for each dataset, and let the model predict the next token as the token with the highest logit. This is not ideal, as the model can even generate tokens which are outside of the space of acceptable answer tokens for the task. This implies that some of our results are even worse than random, as the model can generate tokens that are not even an acceptable response.

This limitation of generating only one token does not allow for techniques that rely on generation of multiple tokens to get an answer, such as \cite{wei2022chain}.

Also, these datasets have no publicly available labels for their test data, so we have used the validation set as our test set.

\section{Finetuning Details}
\label{sec:finetuning}
We do finetuning only in GPT2-XL due to time constraints to get this version of the work, as doing for Phi-2 required us to re-implement the method much more efficiently to fit the model in a single GPU. We use a batch size of 2 (due to memory constraints), learning rate of 0.01 using Adam \cite{kingma2017adam}, and we only train for a single epoch.

\end{document}